\documentclass[11pt,a4paper]{article}
\usepackage[hyperref]{emnlp2018}
\usepackage{times}
\usepackage{latexsym}
\usepackage{graphicx}
\usepackage{comment}
\usepackage{color}
\usepackage{paralist}
\usepackage{xspace}
\usepackage[utf8]{inputenc}
\usepackage{booktabs}
\usepackage{amsmath}
\usepackage{amssymb}
\usepackage{pgfplots}
\pgfplotsset{compat=newest}
\usepackage{url}

\aclfinalcopy 

\title{Learning Concept Abstractness Using Weak Supervision}

\author{
	Ella Rabinovich$^{\blacktriangle}$\thanks{*Work done while the author was at IBM Research.}
	\qquad Benjamin Sznajder$^{\dagger}$ \qquad Artem Spector$^{\dagger}$ \qquad Ilya Shnayderman$^{\dagger}$ \\ 
	\textbf{Ranit Aharonov$^{\dagger}$ \qquad David Konopnicki$^{\dagger}$ \qquad Noam Slonim$^{\dagger}$} \vspace{0.075cm} \\
	$^{\dagger}$IBM Research \\
	$^{\blacktriangle}$ Dept. of Computer Science, University of Toronto \vspace{0.075cm} \\
	\{benjams, artems, ilyashn, ranita, davidko, noams\}@il.ibm.com \\
	ella@cs.toronto.edu
}

\date{}

\begin{document}
\maketitle
\begin{abstract}
We introduce a weakly supervised approach for inferring the property of abstractness of words and expressions in the complete absence of labeled data. Exploiting only minimal linguistic clues and the contextual usage of a concept as manifested in textual data, we train sufficiently powerful classifiers, obtaining high correlation with human labels. The results imply the applicability of this approach to additional properties of concepts, additional languages, and resource-scarce scenarios.
\end{abstract}

\section{Introduction}
\label{sec:intro}
During the last decades, the influence of psycholinguistic properties of words on cognitive processes has become a major topic of scientific inquiry. Among the most studied psycholinguistic attributes are concreteness, familiarity, imagery, and average age of acquisition. \textit{Abstractness} (the opposite of concreteness) quantifies the degree to which an expression denotes an entity that can be directly perceived by human senses.

Word abstractness ratings were first collected by \citet{spreen1966parameters} and \citet{paivio1968concreteness}, and made available in the MRC database \citep{coltheart1981mrc} for 4,292 English words. Since its release, this database has stimulated research in a wide range of linguistic tasks, as well as artificial intelligence and cognitive studies. Despite their evident usefulness, resources providing abstractness ratings are relatively rare and of limited size. Here, we address the task of \textit{automatically} inferring the abstractness rating of a concept by applying a \textit{weakly supervised} approach that exploits minimal linguistic clues.

Studies on derivational morphological processes indicate that word meaning is often entailed by its morphology. As an example, word suffixation by \textit{-ant} or \textit{-ent} is used to denote a person, as in \textit{assistant}, while the suffix \textit{-hood} yields nouns meaning ``condition of being'', as in \textit{childhood}. A wide range of word-formation processes was described by \citet{huddleston2002cambridge}; in particular, the authors detail categories of suffixes that are used to derive words, broadly perceived as abstract, e.g., \textit{-ism} as in \textit{feminism}, or \textit{-ness} as in \textit{agreeableness}.

Concept abstractness indicators are also likely to be manifested in its contextual usage. Consider the two sentences below, each embedding abstract and concrete words – one describing \textit{feminism} and the other \textit{screwdriver} – respectively:

\vspace{0.075cm}
\textit{Second- and third-wave \textbf{feminism} in China involved a reexamination of women`s \underline{roles} during the \underline{communist} \underline{revolution} and other \underline{reform} \underline{movements}, and new discussions about whether women`s \underline{equality} has been fully achieved.}

\vspace{0.075cm}
\textit{Many \textbf{screwdriver} \underline{handles} are not \underline{smooth} and often not \underline{round}, but have \underline{bumps} or other irregularities to improve \underline{grip} and to prevent the \underline{tool} from \underline{rolling} when on a flat \underline{surface}.}
\vspace{0.075cm}

We hypothesize that the immediate neighborhood of a word as reflected in embedding sentences captures the signal of abstractness. In the examples above, several potential clues for the degree of word abstractness are underlined.

Correspondingly, we propose a method for inferring the degree of abstractness of concepts in the complete absence of labeled data, by exploiting (1) a minimal set of morphological word-formation clues; and (2) a text corpus for learning the context in which words tend to appear.

We demonstrate that this method allows us to infer the abstractness ratings of unigram, bigram and trigram Wikipedia concepts (titles) -- the task that, to the best of our knowledge, was only addressed through manual labeling so far \citep{brysbaert2014concreteness}. The main contribution of this work is, therefore, in the proposal and evaluation of a weakly supervised methodology for inferring the abstractness rating of concepts, potentially applicable to additional languages. The suggested approach may also be applicable for predicting other word and concept properties, when those are manifested in both morphology and context. Finally, we release a dataset of 300K Wikipedia concepts automatically rated for their degree of abstractness, and additional 1500 unigram, bigram and trigram concepts annotated with both manual and predicted scores.\footnote{The datasets are available for download at \url{https://www.research.ibm.com/haifa/dept/vst/debating_data.shtml}}

\section{Related work}
A large body of research addressed the relations of word abstractness and cognitive processes \citep{connell2012strength, gianico2012word, oliveira2013roles,nishiyama2013dissociative,paivio2013dual,barber2013concreteness}. Computational investigation of word abstractness and concreteness has been a prolific field of recent research, laying out an empirical foundation for the theoretically motivated hypotheses on the characteristics of these properties. A ranker trained on psycholinguistic features extracted from the MRC database (in combination with other features) reached first place in the English Lexical Simplification task at SemEval 2012 \citep{jauhar2012uow}. \citet{hill2014concreteness} achieved state-of-the-art performance in Semantic Composition and Semantic Modification prediction by including concreteness in the set of features used by the model.

Along the years, several works extended the seed MRC dataset by employing various \textit{supervised} machine learning techniques, further utilizing the extended dataset for tasks of lexical simplification \citep{paetzold2016inferring, paetzold2016collecting}, cross-lingual metaphor detection \citep{tsvetkov2013cross}, literal and metaphorical sense identification \citep{turney2011literal}, as well as readability assessment of Brazilian Portuguese \citep{dos2017lightweight}. \citet{feng2011simulating} exploited word attributes from WordNet, properties extracted from the CELEX database, and Latent Semantic Analysis over a large text corpus for building a linear regression model predicting abstractness rate; the model accounted for 64\% variance of human annotations.

A comprehensive survey of psycholinguistic and memory research on word concreteness is presented in \citet{brysbaert2014concreteness} (BWK), who conducted a large-scale manual annotation of concreteness ratings for over 40K concepts, further used by  \citet{rothe2016ultradense} to infer concreteness ratings for the whole Google News lexicon. To the best of our knowledge, our work is the first attempt to automatically infer the property of concept abstractness in the complete absence of labeled data.

\section{Predicting concept abstractness}
\subsection{Abstractness indicators}
Nominalization is a word-formation process that involves the formation of nouns from bases of other classes by means of affixation. As an example, a derivational suffix can be added to an adjective (capable+\textit{ity} for \textit{capability}) or a verb (react+\textit{tion} for \textit{reaction}) to create a noun. Various word-formation processes often enrich words with meaning associated with certain semantic grouping. \citet{huddleston2002cambridge} detail nominalization processes that serve to form nouns denoting a ``state'' or ``condition of being'', which in turn are broadly associated with abstractness. As such, the suffixes \textit{-ety, -ity} and \textit{\mbox{-ness}} carry over the general meaning of ``quality or state of being'' and the suffix \textit{-ism} is used to form nouns denoting a range of doctrines, beliefs and movements \citep{huddleston2002cambridge}.
Additional suffixes that tend to form English nouns with high degree of abstractness include \textit{-ance, -ence, -ation, -ution, -dom, -hood, -ship} and \textit{-y}.

\subsection{Dataset}
\label{sec:dataset}
We used the English Wikipedia\footnote{We used the Wikipedia May 2017 dump.} article titles as a proxy for retrieving frequently used single- and multi-word expressions, thereby associating over 5M Wikipedia titles with concepts.
\begin{center}
	
\end{center}
\paragraph{Training data}
We chose two abstractness signals, manifested by the suffixes \textit{-ism} and \textit{-ness}, representing different types of abstract meanings. We extracted 1,040 potentially abstract unigram Wikipedia titles suffixed by either of the two (the positive class). The -- admittedly noisy -- concrete (negative) class was generated by randomly selecting the same number of unigram concepts from the complementary set of titles.
In both cases, we set a threshold\footnote{The minimum of 20 occurrences for a concept.} on the frequency of a concept in the corpus, and filtered out non-alphabetic unigrams and unigrams containing special characters. We assessed the quality of the positive and negative weakly-labeled training unigrams by manual annotation of their level of abstractness, obtaining abstractness prior of 93\% in the set of presumably abstract concepts, and concreteness prior of 81\% for the opposite class.

Given this set of weakly-labeled positive and negative concepts, we randomly selected a set of Wikipedia sentences that include any of these concepts (equally split by positive and negative unigrams), to be used in the training phase, while limiting sentence length to the range of 10 to 70 tokens. This step resulted in about 400K train sentences in each class, 800K in total. The final preprocessing phase involved masking a sentence concept with a generic token, aiming to prevent the classifier from training on the concept itself, and instead training on its contextual usage.

\paragraph{Evaluation data}
A randomly selected set of 1500 Wikpedia concepts (with the minimum of 500 occurrences per concept), split equally between unigrams, bigrams and trigrams, and distinct from the training set, was used for testing prediction. We henceforth refer to this set of concepts as the \textit{evaluation set}. Each of these concepts was manually annotated for abstractness on the 1--7 scale by seven in-house labelers, using an adaptation of the guidelines by \citet{spreen1966parameters} to the multi-word scenario:

\textit{Words or phrases may refer to persons, places and things that can be seen, heard, felt, smelled or tasted or to more abstract concepts that cannot be experienced by our senses. The purpose of this task is to rate a list of concepts with respect to "concreteness" in terms of sense-experience. Any expression that refers to objects, materials or persons should receive a high concreteness rating; any expression that refers to an abstract concept that cannot be experienced by the senses should receive a low concreteness rating. Concrete concepts typically have physical or concrete existence, while abstract do not. Think of the concepts "onion" and "nationalism" -- "onion" can be experienced by our senses and therefore should be rated as concrete (1); "nationalism" cannot be experienced by the senses as such and therefore should be rated as abstract (7).}

Word polysemy is a common challenge in tasks related to lexical semantics. As such, our perception of the concreteness rate of the concept \textit{bank} may vary depending on whether a financial institution or a river bank is concerned. While we could not avoid this issue altogether (since working with pre-trained word representations that do not carry disambiguation information), we ensured that all in-house labelers annotated the same word sense by providing them with Wikipedia definition of the most frequent sense of a concept.

The final abstractness score was computed as the average over individual annotations. The average pairwise weighted Kappa agreement\footnote{We used the implementation in \url{http://scikit-learn.org}, with ``quadratic'' scheme.} on the entire set of 1500 concepts was 0.65.

\subsection{Classification models}
\label{sec:classifiers}
We hypothesize that words that share similar degree of abstractness tend to share certain similarities in their contextual usage; that, in contrast to concepts that exhibit opposite abstractness rate. Indeed, a statistical significance test applied to the (weak) positive and negative training data (Section \ref{sec:dataset}) reveals markers such as \{\textit{parish, movement, century, spiritual, life, doctrine, nature, regime}\} sharing excessive frequency in sentences containing abstract concepts. The very essence of this phenomenon is captured by distributed word representations \citep{mikolov2013distributed, pennington2014glove}, a.k.a.\ word embeddings, learned based on the contextual usage of words. We therefore trained three classifiers, each exploiting different language properties, as described below.

\paragraph{Naive Bayes (NB)}
Using solely word counts in textual data, we used a simple probabilistic Naive Bayes classifier, with a bag-of-words feature set extracted from the 800K sentences containing positive and negative training concepts. Given a sentence containing a test concept, its degree of abstractness was defined as the posterior probability assigned by the classifier. Aiming at robust classification, we retrieved 500 sentences containing each test concept from the corpus. Consequently, the final abstractness score of a concept was calculated by averaging the predictions assigned by the classifier to individual sentences.

\paragraph{Nearest neighbor}
We used the nearest neighbors algorithm, specifically, its \textit{radius-based} version (NN-RAD), using the pre-trained GloVe embeddings \citep{pennington2014glove}. This classifier estimates the degree of concept abstractness given only its distributional representation.

The abstractness score of a test concept was computed by the ratio of its abstract neighbors to the total number of concepts within the predefined radius, where the entire set of neighbors is limited to the concepts in the weakly-labeled training set. The proximity threshold (radius) was set to 0.25, w.r.t. the cosine similarity between two embedding vectors.\footnote{The radius was tuned on the set of 500 unigrams.} Multi-word concepts were subject to more careful processing, where the classifier computed a multi-word concept representation as an average of representations of its individual words, and further estimated the abstractness score of the obtained embedding. In case that one of a concept constituents was not found in embeddings, we excluded the concept from computation.

\paragraph{RNN}
Aiming at exploiting both embeddings and textual data, we utilized a bidirectional recurrent neural network (RNN) with one layer of forward and backward LSTM cells. Each cell has width of 128, and is wrapped by a dropout wrapper with keep probability 0.85. An attention layer was created in order to weigh words according to their proximity to the train/test concept. The output of the LSTM cells is passed to the attention layer which reduces it to the size of 100. The output of the attention layer is passed to a fully connected layer which produces the final prediction of the abstractness level of a concept. GloVe embeddings with 300 dimensions were used as word representations. Given a set of sentences containing a test concept, its final abstractness score was computed by applying the averaging procedure described for the Naive Bayes classifier.

\section{Results}
We demonstrate that trained models discover linguistic patterns associated with abstract meaning (beyond those known at training), and furthermore yield abstractness scores that correlate significantly with human annotations.

\subsection{Revealing abstractness markers}
We automatically scored 100K unigram Wikipedia concepts for abstractness with all classifiers and extracted the set of suffixes that share excessive frequency in the top-k abstract concepts using the statistical proportion test. More specifically, we applied the test to the exhaustive list of all three-character English suffixes (e.g., \textit{\mbox{-aaa}, \mbox{-aab}}), counting their occurrences in the subset of concepts with the highest abstractness scores\footnote{We used the set of 18\% highest ranked concepts -- the fraction of abstract concepts in a sample population, as estimated by manual labeling.} (the population under test) and in the remainder (the background). Our hypothesis was that suffixes associated with abstract meaning in the literature will be over-represented in the population of concepts ranked as abstract by the classifiers. The top-10 suffixes, scored by their statistical significance p-value\footnote{In all cases the obtained p-value was practically zero.} were \{\textit{\textbf{-ism}, -ity, -ion, -sis, -ics, \textbf{-ess}, -phy, -nce, -ogy, -ing}\} -- suffixes broadly associated with abstractness in the literature (where all suffixes but two are distinct from the training data). The underlying concept examples included \{\textit{illegalism, modernity, antireligion, henosis, politics, lawlessness, ecosophy, conscience, ideology, enabling}\} -- words broadly perceived as abstract.

\subsection{Abstractness rating}
\label{sec:rating}
Table \ref{tbl:examples} presents a few examples of abstract and concrete concepts, as identified by manual annotation, along with their abstractness score as predicted by the RNN classifier (Section~\ref{sec:classifiers}).

\begin{table}[hbt]
\centering
\resizebox{\columnwidth}{!}{
\begin{tabular}{l|c|l|c}
\multicolumn{2}{c|}{abstract} & \multicolumn{2}{c}{concrete} \\ \hline
concept & score & concept & score \\ \hline
marxism & 0.972 & plywood & 0.000 \\
islamophobia & 0.969 & Wiltshire & 0.000 \\
affirmative action & 0.844 & moonlight & 0.058 \\
absolute monarchy & 0.842 & convoy & 0.112 \\
sincerity & 0.836 & gadget & 0.120 \\
\end{tabular}
}
\caption{\small{Examples of concepts found as abstract/concrete (above/below the average score of 0.5) via manual annotation, along with their score as predicted by RNN.}}
\label{tbl:examples}
\end{table}

Table \ref{tbl:correlation} presents the Pearson correlation between the abstractness scores as assigned by the classifiers and the manual annotations over the evaluation set. We also present the correlation of scores produced by our classifiers to the set of Wikipedia concepts from the manually annotated MRC database (MRC-seed, Section \ref{sec:intro}), and to the set of 5883 noun concepts\footnote{Only concepts that can be mapped to a corresponding Wikipedia page were considered.} from manually annotated BWK dataset \citep{brysbaert2014concreteness}.

\begin{table}[hbt]
\centering
\resizebox{\columnwidth}{!}{
\begin{tabular}{l|c|c|c}
\multicolumn{1}{c|}{{test set}} & \multicolumn{1}{c|}{{Naive Bayes}} & \multicolumn{1}{c|}{{NN-RAD}} & \multicolumn{1}{c}{{RNN}} \\ \hline
BWK & \textbf{0.657} & 0.622 & 0.634 \\ \hline
MRC-seed & \textbf{0.674} & 0.576 & 0.669 \\ \hline
1-grams & 0.679 & 0.638 & \textbf{0.740} \\
2-grams & 0.565 & 0.515 & \textbf{0.666} \\
3-grams & 0.412 & 0.467 & \textbf{0.490} \\
\end{tabular}
}
\caption{\small{Correlation of abstractness scores assigned by the classifiers to manual annotations.}}
\label{tbl:correlation}
\end{table}

Evidently, the best results are obtained by the RNN classifier, yielding up to 0.740 correlation with human annotations. Notably, the simple Naive Bayes, utilizing only textual data, yields results of reasonable quality; the broad implication of this outcome lies in the potential applicability of this approach to resource-scarce scenarios where high quality word embeddings are not available. Interestingly, while using Google word2vec embeddings (instead of Glove) yielded similar results, utilizing fastText pre-trained representations \citep{joulin2016bag} obtained more accurate ranking, e.g., the NN-RAD classifier yielded correlation of 0.688 for the BWK dataset, compared to 0.622 obtained using Glove (Table~\ref{tbl:correlation}). We attribute this improvement to the fact that fastText embeddings better capture morphological word properties and cover more extensive vocabulary. 

The relatively low correlation obtained with trigram concepts can be explained by the inherent complexity introduced by the multi-word scenario, challenging still further the subjective human perception of abstractness. While inter-labeler agreement for unigrams and bigrams was 0.72 and 0.66, respectively, it only reached 0.54 for trigrams, supporting the aforementioned hypothesis. 

\subsection{Varying the size of a test set}
How many sentences containing a test concept suffice for a reliable prediction? We address this question by limiting the number of (randomly chosen) sentences used for rating. While the correlation obtained by RNN with 500 sentences containing a test concept reached 0.740 (Table~\ref{tbl:correlation}), as little as 10, and even 5 sentences yielded correlation of 0.706 and 0.675, respectively, implying the efficiency and effectiveness of the presented approach in the availability of only little data. The plot in Figure~\ref{fig:sensitivity} presents the correlation of the RNN and NB classifiers to label as function of number of (randomly sampled) sentences used for evaluation. Each such experiment (e.g., using 1, 5, 10 sentences) was averaged over 50 runs; the average correlation to label, as well as standard deviation, are plotted on the chart. The constant correlation yield by the (text-independent) NN-RAD algorithm is illustrated by the vertical line.

\pgfplotsset{every axis/.append style={
		label style={font=\small},
		tick label style={font=\small}
}}

\usepgfplotslibrary{fillbetween}

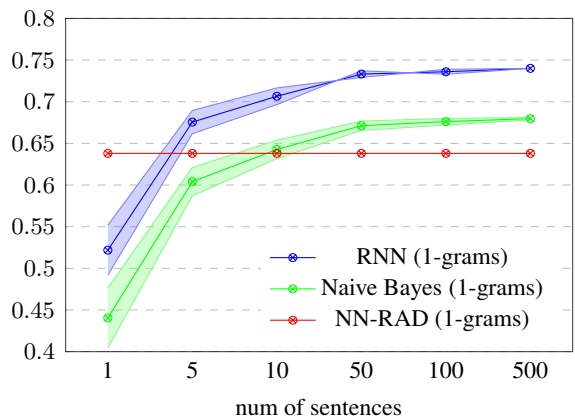
\begin{figure}[hbt]
	\centering
	\begin{tikzpicture}
	\begin{axis}[ymin=0.4, ymax=0.8, xmin=1, xmax=6,
	ytick={0.4,0.45,0.5, 0.55, 0.6, 0.65, 0.7, 0.75,0.8}, ytick align=inside, ytick pos=left,
	xtick={1, 2, 3, 4, 5, 6}, xtick align=inside, xtick pos=left,
	xticklabels = {1, 5, 10, 50, 100, 500},
	tick align=inside,
	tickwidth=0pt,
	y axis line style={opacity=0.75},
	enlarge x limits=true,
	xlabel=num of sentences,
	legend pos=south east,
	legend style={draw=none, font=\small},
	ymajorgrids=true,
	grid style=dashed,
	height=6cm,
	width=8.25cm,
	]
	
	\addplot[blue, mark=otimes, mark options={blue, scale=0.75}] coordinates {(1,0.5218) (2, 0.6756) (3, 0.7066) (4, 0.7332) (5, 0.7360) (6, 0.7401)};
	\addplot[name path=rnn_top,color=blue!50] coordinates {(1,0.5518) (2, 0.6896) (3, 0.7166) (4, 0.7292) (5, 0.7390) (6, 0.7401)};
	\addplot[name path=rnn_down,color=blue!50] coordinates {(1,0.4918) (2, 0.6616) (3, 0.6966) (4, 0.7372) (5, 0.7330) (6, 0.7401)};
	\addplot[blue!50,fill opacity=0.35] fill between[of=rnn_top and rnn_down];
	
	\addplot[green, mark=otimes, mark options={green, scale=0.75}] coordinates {(1, 0.4405) (2, 0.6041) (3, 0.6427) (4, 0.6712) (5, 0.6762) (6, 0.6797)};
	\addplot[name path=rnn_top,color=green!50] coordinates {(1, 0.4765) (2, 0.6211) (3, 0.6537) (4, 0.6768) (5, 0.6800) (6, 0.6810)};
	\addplot[name path=rnn_down,color=green!50] coordinates {(1, 0.4045) (2, 0.5871) (3, 0.6317) (4, 0.6656) (5, 0.6720) (6, 0.6783)};
	\addplot[green!50,fill opacity=0.35] fill between[of=rnn_top and rnn_down];
	
	\addplot[red, mark=otimes, mark options={red, scale=0.75}] coordinates {(1, 0.6380) (2, 0.6380) (3, 0.6380) (4, 0.6380) (5, 0.6380) (6, 0.6380)};
	\legend{RNN (1-grams),,,, Naive Bayes (1-grams),,,, NN-RAD (1-grams)}
	
	\end{axis}
	\end{tikzpicture}
	\label{fig:correlation}
	\vspace{-0.25em}
	\caption{Average correlation (and standard deviation) to manual annotation as function of number of sentences used for evaluation.}
	\label{fig:sensitivity}
\end{figure}

\subsection{Comparison to supervised models}
\citet{tsvetkov2013cross} used supervised learning algorithm to propagate abstractness scores to words using pre-trained word representations. Utilizing vector elements as features, they trained a supervised classifier, and predicted the degree of abstractness for unseen words. Abstractness rankings from the MRC database were used as a training set, and the classifier predictions were binarized into abstract-concrete boolean indicators using predefined thresholds. The authors obtained 94\% accuracy when tested on held-out data.

\section{Conclusions}
We presented a weakly supervised approach for inferring the degree of concept abstractness. Our results demonstrate that a minimal morphological signal and a textual corpus are sufficient to train classifiers that yield relatively accurate predictions, that in turn can be used to unravel additional linguistic patterns indicative of the same property. Our future plans include exploring the value of the proposed methodology with other languages and additional properties.

\section*{Acknowledgments}
We are grateful to Dafna Sheinwald and Shuly Wintner for much advise and helpful suggestions. We also thank our anonymous reviewers for their constructive feedback.

\bibliography{abstractness}
\bibliographystyle{acl_natbib_nourl}

\end{document}